\documentclass[letterpaper]{article} 
\usepackage{aaai2026}  
\usepackage{times}  
\usepackage{helvet}  
\usepackage{courier}  
\usepackage[hyphens]{url}  
\usepackage{graphicx} 
\usepackage{natbib}  
\usepackage{caption} 
\usepackage{algorithm}
\usepackage{algorithmic}
\usepackage{booktabs, array, tabularx}
\usepackage{rotating}
\usepackage{threeparttable}
\usepackage{float}
\usepackage[symbol]{footmisc}
\usepackage{tikz}
\usetikzlibrary{calc}
\usepackage{newfloat}

\usepackage{listings}
\DeclareCaptionStyle{ruled}{labelfont=normalfont,labelsep=colon,strut=off} 
\floatstyle{ruled}
\newfloat{listing}{tb}{lst}{}
\floatname{listing}{Listing}

\title{FinForge: Semi-Synthetic Financial Benchmark Generation}
\author {
    Glenn Matlin\textsuperscript{\rm 1 2 3},
    Akhil Theerthala\textsuperscript{\rm 1 \footnotemark[1]},
    Anant Gupta\textsuperscript{\rm 3 \thanks{Equal Contribution}},
    Anirudh Jaidev Mahesh\textsuperscript{\rm 2},
    Yi Mei Ng\textsuperscript{\rm 2},
    Rayan Castilla\textsuperscript{\rm 3},
    Sudheer Chava\textsuperscript{\rm 1 2 3}
}
\affiliations {
    \textsuperscript{\rm 1}Financial Services Innovation Lab, Georgia Institute of Technology\\
    \textsuperscript{\rm 2}College of Business, Georgia Institute of Technology\\
    \textsuperscript{\rm 3}College of Computing, Georgia Institute of Technology\\
    glenn@gatech.edu, akhiltvsn@gmail.com, agupta886@gatech.edu
}

\usepackage{bibentry}

\usepackage{xspace}      
\usepackage{xcolor}      
\usepackage{pifont}      
\usepackage{enumitem}    

\newcommand{\eg}{e.g.,\xspace}      

\newif\ifdraft
\drafttrue  

\begin{document}

\maketitle

\begin{tikzpicture}[remember picture, overlay]
    \node[anchor=south, inner sep=1.5cm] at ($(current page.south)+(0, 1.2cm)$) {
        \begin{minipage}{0.95\textwidth} 
            \centering
            \includegraphics[width=0.58\paperwidth]{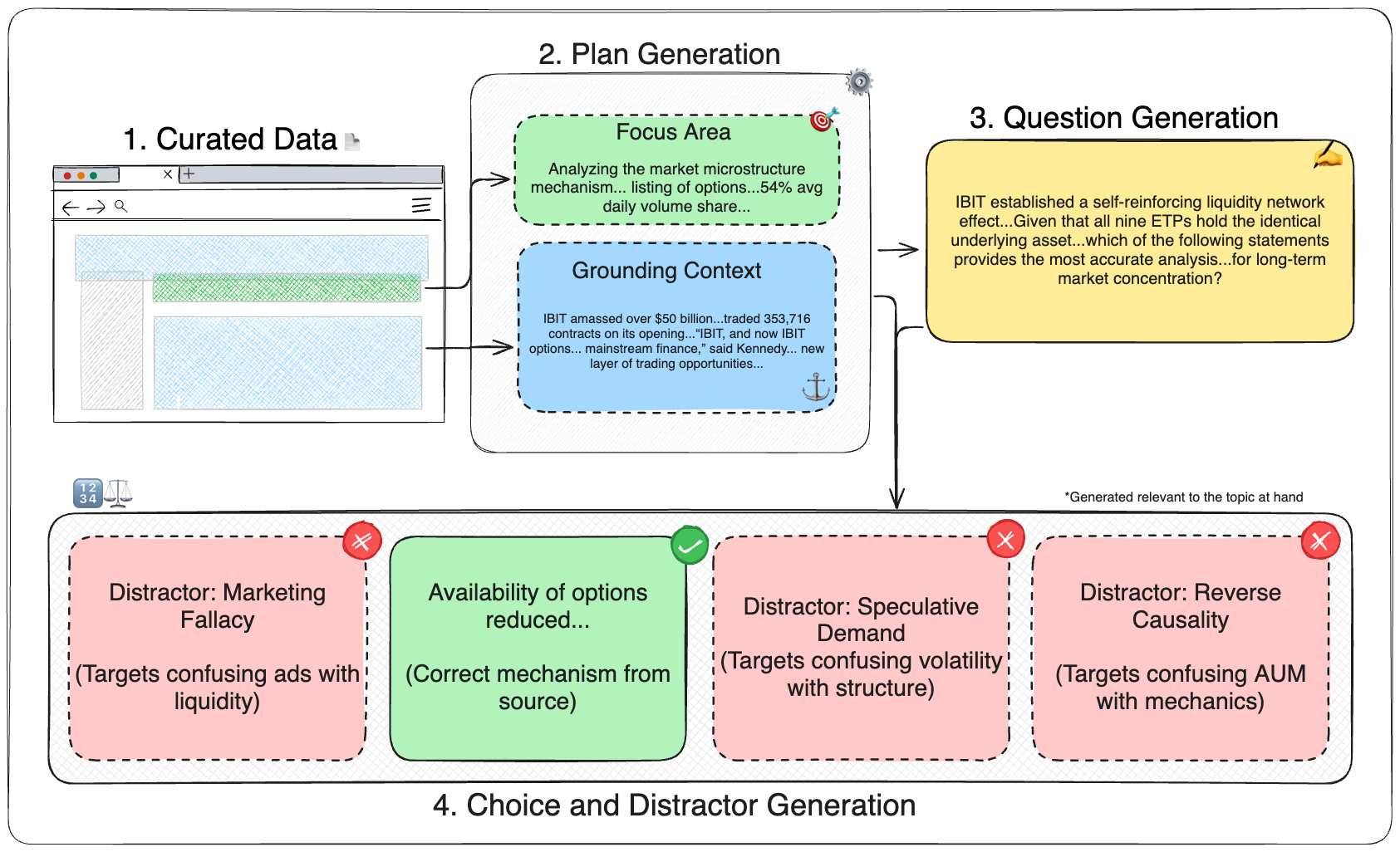}
            
            \captionof{figure}{The FinForge Process - The framework ingests raw unstructured financial text (1) and identifies explicit causal triggers and outcomes. Crucially, the Reasoning Engine (2) applies domain-specific schemas to infer the implicit financial mechanism linking them. Finally, it synthesizes a complex, multiple-choice question and answer (3).}
            \label{fig:finforge_example}
        \end{minipage}
    };
\end{tikzpicture}

\enlargethispage{-9cm}

\begin{abstract}
Evaluating Language Models (LMs) in specialized, high-stakes domains such as finance remains a significant challenge due to the scarcity of open, high-quality, and domain-specific datasets. Existing general-purpose benchmarks provide broad coverage but lack the depth and domain fidelity needed to assess LMs' capabilities for real-world financial reasoning, which requires both conceptual understanding and quantitative rigor. To address this gap, we introduce FinForge, a scalable, semi-synthetic pipeline for constructing finance-specific evaluation benchmarks through a hybrid of expert-guided data curation and controlled LM-based synthesis. FinForge combines manual and programmatic corpus construction from authoritative financial sources with structured question generation and validation using \texttt{Gemini~2.5~Flash}. To demonstrate the pipeline's efficacy, we produce FinForge-5k, a snapshot benchmark comprising over 5{,}000 human-validated question–answer pairs across 11 finance subdomains, derived from a curated corpus of 100{,}000 verified documents totaling 143M tokens. Evaluation of state-of-the-art open-source and closed-source models on FinForge-5k reveals significant differences in financial reasoning, with leading models achieving accuracy levels near 80\%. These findings underscore the framework's utility for diagnosing current model limitations and guiding future improvements in financial domain competence. All code and data are available at \texttt{https://github.com/gtfintechlab/FinForge}.

\end{abstract}


\enlargethispage{-9cm}

\section{Introduction}
Language Models (LMs) are increasingly adopted for decision support in complex, high-stakes domains such as finance, law, and public policy \cite{bommasani2022opportunities,finllms2024survey}. While recent advances in LMs have demonstrated strong performance on general knowledge benchmarks \cite{hendryckstest2021} and professional exams \cite{openai2023gpt4}, reliably evaluating these systems in specialized, knowledge-intensive, and dynamic domains remains a significant challenge. Existing evaluation sets offer broad subject coverage and serve as useful indicators of general knowledge. However, static benchmarks are limited by potential data leakage into LM training corpora \cite{deng2024investigating}, which can artificially inflate performance due to LM memorization. These challenges are exacerbated in dynamic domains like finance, where knowledge must be continually updated to reflect real-world developments. Empirical evidence suggests that top-performing general models do not necessarily excel at financial tasks requiring domain-specific nuance or quantitative reasoning \cite{chen2021finqa,Islam2023FinanceBenchAN}. This highlights the necessity of dynamic benchmark generation to rigorously assess LM financial knowledge and reasoning robustness in realistic industry scenarios.

Finance presents unique challenges for LM evaluation due to its multi-domain complexity, stochastic characteristics, and stringent regulatory requirements \cite{ai2022finance,ai2025integration,dllf2024survey,biblio2024ai}. Effective financial analysis requires both comprehensive domain knowledge---such as familiarity with financial instruments, regulations, and policies---and advanced quantitative problem-solving using real-world data, including asset valuations and risk projections. Additionally, the financial sector evolves rapidly, with new markets, regulations, and trends emerging continuously due to the dynamic nature of global economic systems. Consequently, maintaining up-to-date knowledge is essential for accurate financial reasoning. These challenges motivate three research questions:
\begin{itemize}
    \item \textbf{RQ1:} Can semi-synthetic benchmark generation produce high-quality, contamination-free evaluation datasets for specialized domains like finance?
    \item \textbf{RQ2:} How do state-of-the-art language models perform across financial subdomains, and what patterns emerge?
    \item \textbf{RQ3:} What reasoning capabilities---conceptual versus quantitative---do current models lack in financial contexts?
\end{itemize}

To address these questions, this paper introduces the FinForge framework, a methodology for generating semi-synthetic benchmarks tailored to the financial domain. FinForge addresses the gap in the dynamic evaluation of language models' financial knowledge by providing a novel pipeline to generate diverse, challenging finance questions grounded in real-world content on demand. The methodology combines human-guided data curation with LM-driven question synthesis, thereby overcoming the limitations of prior static, easily memorizable benchmarks. By leveraging a hybrid workflow, FinForge allows for the continuous creation of contamination-free evaluation sets that evolve alongside the dynamic financial landscape.

The FinForge methodology curates relevant knowledge exclusively from authoritative sources, such as academic textbooks, institutional research, and domain experts, deliberately excluding user forums and trivial stock data. By leveraging verified knowledge, the pipeline increases the difficulty of AI evaluations by generating challenging scenarios that test domain-specific problem-solving capabilities. The process employs a multi-stage language model workflow to analyze long-context documents, extract key information, and generate knowledge for creating question–answer pairs. Each question is planned by identifying a central concept or reasoning challenge within the source material, formulating a complex question with embedded background information, and providing a correct answer and distractors. A more advanced language model serves as a judge to validate the quality and financial relevance of each question. This controlled generation process produces difficult, self-contained questions that require expert-level economic insight and multi-step reasoning.

To demonstrate the framework's utility, we generate FinForge-5k, a snapshot benchmark comprising 5{,}000 expert-level finance question–answer pairs. This benchmark is derived from a high-quality corpus of up-to-date financial documents—curated via the pipeline—totaling 143M tokens across more than 100{,}000 verified articles spanning 11 subdomains, including personal finance, corporate finance, macroeconomics, and securitized investments. Notably, the underlying methodology is dynamic: the FinForge framework can be rerun and new documents incorporated to update the question set, allowing continual adaptation to emerging financial knowledge. This paper details the FinForge methodology and demonstrates its capabilities by benchmarking several state-of-the-art models on FinForge-5k, thereby highlighting current strengths and weaknesses in financial reasoning.

\begin{figure*}[h!]
    \centering
    \includegraphics[width=0.8\textwidth]{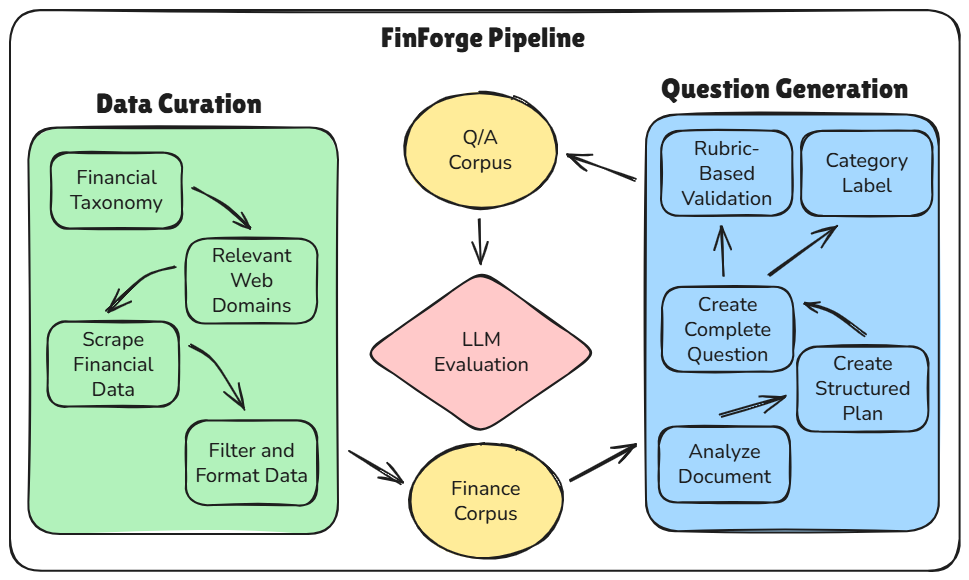}
    \caption{The FinForge pipeline comprises two complementary stages. \textbf{Data Curation} (left): A hybrid manual--programmatic process that applies a financial taxonomy to identify authoritative web domains, scrapes and filters content, and assembles a high-quality Finance Corpus. \textbf{Question Generation} (right): An LM-driven five-stage workflow that analyzes documents to extract salient information, creates structured answer plans, generates self-contained questions with plausible distractors, assigns category labels, and applies rubric-based validation to ensure relevance, clarity, and factual accuracy. The validated outputs populate both a Q/A Corpus for benchmarking and feed back into the Finance Corpus for iterative refinement.}
    \label{fig:qagen}
\end{figure*}

\section{Related Works}

Existing public datasets for evaluating Question Answering (QA) in the financial domain are limited in scope, focus, and recency. While several benchmarks have made significant contributions to specific subtasks, their coverage remains narrow. For example, FinQA \cite{chen2021finqa} and its conversational extension ConvFinQA \cite{chen2022convfinqa} provide 8,200 QA pairs centered on numerical reasoning with plain-text representations of tabular financial data, and TAT-QA \cite{zhu-etal-2021-tat-qa} similarly targets QA on annual reports with tables and text. Although valuable, these datasets primarily address numeric reasoning and do not encompass broader conceptual finance knowledge. The recent FinanceBench \cite{Islam2023FinanceBenchAN} initiative sought to expand the range of finance QA but offers only 150 question–answer pairs, with relatively simple questions that do not reflect real-world complexity. Similarly, FinTextQA, a long-form QA dataset compiled from textbooks and policy documents, encourages explanatory answers and highlights the limitations of prior benchmarks that focused predominantly on stock data or basic calculations. In the industry, the scarcity of robust evaluation sets has prompted efforts such as S\&P Global Kensho's S\&P AI Benchmarks, which assembled 600 expert-verified questions across categories, including domain knowledge, quantity extraction, and quantitative reasoning. The emergence of these initiatives reflects the growing demand for more realistic and challenging evaluations of LMs in finance.

A central challenge in constructing dynamic benchmarks is ensuring that questions remain both novel and sufficiently difficult. Since most language models are trained on extensive public internet data, static test suites risk being memorized during training, thereby compromising their evaluative value. To address this, researchers are increasingly exploring dynamic or semi-synthetic benchmarks that can be refreshed or generated as needed \cite{Das2021AutomaticQG, Guo2024SurveyNQG}. Previous studies have demonstrated that carefully controlled LM-based generation can yield high-quality, semi-synthetic data at scale, enabling contamination-free evaluation \cite{long2024llmsyntheticsurvey}.

\subsection{General Knowledge and QA Benchmarks}

Several benchmarks have been developed to measure broad knowledge and reasoning ability in LMs. MMLU \cite{hendryckstest2021}, which tests models on everything from elementary math to professional law and accounting, has become a standard evaluation suite. Other efforts include Big-Bench \cite{bigbench} and the AI2 Reasoning Challenge (ARC) \cite{allenai:arc,arcbench}, which target complex reasoning or scientific questions. These static benchmarks have driven progress, but are increasingly prone to contamination from training data as models ingest questions and answers from the web \cite{benchmark2024contamination,xu2024benchmarking}. This has sparked interest in more adaptive evaluation methods.

One line of work uses LM-based generation to create new test items, prompting LMs to rewrite or expand existing benchmark questions into novel variants. More holistically, LatestEval \cite{li2024latesteval} constructs entirely new reading comprehension sets from real-time sources such as BBC News, using an LM to generate questions for up-to-date passages.

Recently, multi-agent systems have been proposed for automated benchmark creation: BenchAgents \cite{butt2024benchagents} splits the task into planning, generation, and verification agents that collaborate (with humans in the loop) to produce high-quality evaluation data. Such approaches yield dynamically extendable benchmarks and help ensure test data novelty. In parallel, research on controllable question generation has introduced techniques to enforce difficulty and content constraints on generated questions. Notably, \citet{pfqs} propose a ``Planning First, Question Second'' (PFQS) method in which an LM first outlines a detailed answer plan (with target answer, relevant facts, and cognitive steps), and then another LM generates a question conforming to this plan. This leads to more faithful, expert-aligned questions, as the model must adhere to a blueprint for the desired reasoning. Our FinForge methodology draws inspiration from these advances in controllable question generation.

\subsection{Financial QA Datasets and Benchmarks}

Before our work, relatively few datasets existed for evaluating QA or reasoning in the financial domain, and each covered only a slice of the domain. FinQA was one of the first, featuring questions that require numerical reasoning over company financial reports. It introduced the challenge of performing multi-step arithmetic operations on statements and tables, and showed that models lag far behind human experts on such tasks. TAT-QA (Table-and-Text QA) similarly targets reasoning with hybrid data (earnings tables + text) in financial reports. These datasets primarily evaluate the ability to do structured data reasoning (\eg reading an annual report) and include annotated programs or formulas for interpretability. A later extension, ConvFinQA, turned FinQA into a multi-turn dialogue challenge. Beyond corporate reports, other benchmarks have been even more limited: the FiQA challenge \cite{Maia2018FiQA} released a small set of user-submitted questions and answers on personal finance topics. Most of these lack the complexity and diversity of knowledge needed to test an advanced AI's full financial acumen. For instance, FinQA and TAT-QA do not include conceptual questions on economics or open-ended advisory questions, focusing instead on factoid numeric problems.

A recent effort, FinanceBench \cite{Islam2023FinanceBenchAN}, sought to compile a wider range of financial QA pairs (covering banking, markets, accounting, etc.). Still, it contains only 150 questions in total – too small to capture the breadth of finance or to reliably benchmark modern LMs. Moreover, FinanceBench's question complexity remains limited, falling short of the complexity of real-world expert queries. Recognizing these gaps, Chen et al. introduced FinTextQA \cite{chen-etal-2024-fintextqa}, a long-form QA dataset drawn from finance textbooks and government agency documents. FinTextQA's 1,262 questions are designed to elicit paragraph-length answers, emphasizing explanatory responses over simple calculations. This provides a valuable test of explanation and retrieval capabilities. However, the dataset is relatively small and requires generative answers rather than the multiple-choice format often used in benchmarking. Complementary to academic datasets, industry researchers have also developed proprietary benchmarks. Notably, Kensho's S\&P ``BizBench'' (developed by Koncel-Kedziorski et al.) evaluates models on finance and business tasks across three main categories: domain knowledge (\eg concept definitions or CFA exam questions), quantitative reasoning (multi-step problems requiring math and finance formulas), and quantity extraction from financial documents. Recent studies have specifically evaluated LLMs on CFA examinations, revealing significant gaps in professional-level financial reasoning \cite{cfa2024evaluation,cfa2025advanced}. The benchmark consists of 600 expert-curated questions and includes eight tasks for financial reasoning, ranging from code generation for math problems to the FinKnow QA task for conceptual questions. A public leaderboard shows that even top-tier models (\eg Claude 3.5, GPT-4) struggle with the hardest numerical reasoning questions \cite{koncelkedziorski2024bizbench}. These efforts emphasize the growing importance of domain-specific evaluation. Our work differs in that we propose an open-source, automated pipeline to generate a much larger set of finance QA pairs (5,000), blending the breadth of coverage of FinanceBench/FinTextQA~\cite{Islam2023FinanceBenchAN} with the realism and difficulty seen in expert-written exams. FinForge's use of real financial texts as the grounding for each question ensures that the content is current and verified, addressing both the dynamic knowledge aspect and the quality-control issue by providing source evidence for each answer. We view FinForge as a complement to prior benchmarks – pushing the envelope on scale and difficulty – and hope it will enable more robust assessment and improvement of LMs for financial applications.

\section{Methodology}

To enable scalable and controlled evaluation of financial reasoning, we required a robust corpus of finance-relevant documents encompassing both numerical and conceptual knowledge. Specifically, the corpus needed to capture (i) quantitative material such as financial calculations and analytical exercises, and (ii) qualitative content reflecting economic principles, market behavior, and institutional context. Given the scarcity of open-source datasets with verified, diverse, and high-quality financial text, we constructed a semi-synthetic dataset using a two-stage pipeline that integrates expert-guided data curation and LM-based question generation. In the first stage, we curated a corpus of high-quality financial documents drawn from authoritative web sources, leveraging a hybrid manual–programmatic pipeline that combines domain expertise with automated filtering and extraction. In the second stage, we employed frontier LMs (specifically, \texttt{Gemini 2.5 Flash})\cite{gemini} to generate diverse, high-quality question–answer pairs from a representative subset of this corpus. Together, these stages yield a scalable, high-fidelity foundation for benchmarking and training models on financial reasoning tasks.

\subsection{Data Curation}
We structured our data curation methodology as a hybrid manual--programmatic pipeline that balances domain expertise with scalability. To minimize effort while maximizing domain coverage and quality, we first decomposed finance into a structured hierarchy of 11 subdomains, including personal finance, corporate finance, investment theory, and macroeconomics. This taxonomy was guided in part by authoritative educational frameworks to ensure conceptual completeness and internal consistency.

For each subdomain, humans identified authoritative web domains based on content rigor, institutional credibility, and topical relevance. Sources without clear editorial oversight or academic grounding---such as discussion forums or informal opinion sites---were systematically excluded. This filtering ensured that only high-quality, verifiable content was included in the corpus.

The pipeline then applies a suite of open-source tools and heuristics---including domain whitelisting, keyword co-occurrence, sitemap traversal, and link structure analysis---to automatically identify, filter, and rank candidate sites for extraction. Once candidate sites were finalized, we leveraged each site's structure to efficiently extract relevant financial text, avoiding the need for exhaustive manual traversal. For text extraction and parsing, we employed \texttt{Trafilatura}\footnote{\url{https://trafilatura.readthedocs.io/}} and \texttt{BeautifulSoup}\footnote{\url{https://beautiful-soup-4.readthedocs.io/}} for HTML-based content, and \texttt{PyMuPDF4LLM}\footnote{\url{https://pymupdf.readthedocs.io/}} for PDF documents, ensuring consistent text normalization and formatting across source types.

The process cleanly separates filtering and extraction stages, enabling parallelized domain filtering and asynchronous content extraction at scale.

\subsection{Question Generation and Validation}
Recent works in question generation demonstrate the scalability and controllability of using tailored LM agents for this task \cite{pfqs, savaal}. Our approach synthesizes insights from these methods and adapts them for controlled generation from domain-specific documents.

FinForge employs an automated five-stage process, as illustrated in Figure~\ref{fig:qagen}. First, we perform a deep analysis of the input documents to extract salient information. This analysis informs the generation of a structured answer plan, which serves as a blueprint for guiding question formulation. Using this plan, we then generate complex questions that remain strictly grounded in the source material. In the final stage, an LM-as-a-judge framework validates these questions against a predefined rubric, filtering for relevance and quality.

\subsubsection{Synthetic Question Generation}
The preliminary document analysis phase aims to discern deep financial thinking patterns within articles to uncover opportunities for probing questions. We specifically focus on identifying four essential characteristics in a document: causal relationships, prominent and competing hypotheses, necessary assumptions, and counterfactual possibilities.

This breakdown of the document is crucial during the planning stage. In this stage, we translate the unstructured information into a concrete blueprint for generation. This involves identifying a specific, testable conceptual nucleus within the text that serves as the ``focus area.'' The agent also assesses the cognitive complexity of this concept, assigning a difficulty rating on a five-point scale ranging from basic recall (1) through multi-step reasoning (3) to expert-level synthesis requiring multiple constraints (5). Finally, it extracts the minimal set of relevant passages required to construct a self-contained question. This plan ensures that the question is well-defined, appropriately challenging, and directly traceable to the source document.

In the third stage, we use this blueprint to formulate a complete question–answer pair. Adhering to the principle of self-containment, all necessary context and data from the relevant passages are embedded directly into the question's premise. The language model for this stage is prompted to generate a natural-language question, plausible distractors, and a concise explanation for the correct answer, all while adhering to the domain-specific requirements provided in the inputs.

In addition to the preceding processes, we implement a supplementary labeling phase to categorize the artifacts based on the financial issue, perceived difficulty, and the targeted model's finance-related capability. These labels help filter out irrelevant questions and provide valuable insights into a model's performance across diverse settings. Each label is linked to its own case-specific rules that help the model adapt to the nature of the topic it must address.

\subsubsection{Validation and Filtering}
The primary issue of automated generation is that outputs frequently fail to meet benchmark standards. We frequently observed unclear inquiries, erroneous hypotheses, or entirely unrelated questions arising from ostensibly direct materials. Consequently, we implemented an extra question validation phase that uses a language model to assess the question's validity for the specific use case. Each question is evaluated across multiple dimensions---financial relevance (domain appropriateness), self-sufficiency (answerable without external context), logical consistency (no contradictions), clarity (unambiguous wording), and complexity (appropriate difficulty). A question passes validation only if it satisfies all five criteria; failure on any dimension results in rejection.

Each iteration of the pipeline development has undergone rigorous validation through expert review, independent of the existing automated checks. Based on this expert feedback, we iteratively refined the generation and filtering logic to produce a final set of higher-quality, more challenging questions.

\begin{table*}[ht!]
\centering
\small
\setlength{\tabcolsep}{3pt}
\renewcommand{\arraystretch}{1.1}

\begin{tabular}{l|r|rrrrrrrrrrr}
\hline
\textbf{Models} & \textbf{Overall} &
\shortstack[c]{Alt/RE} &
\shortstack[c]{Beh/Quant} &
\shortstack[c]{Corp Fin\\\& Val} &
\shortstack[c]{FinTech} &
\shortstack[c]{FAR} &
\shortstack[c]{Ethics \&\\Gov} &
\shortstack[c]{Mkt \&\\Deriv} &
\shortstack[c]{Reg \&\\Comp} &
\shortstack[c]{Portf\\Mgmt} &
\shortstack[c]{Wealth\\Mgmt} &
\shortstack[c]{Pub/Intl\\Fin} \\
\hline
Qwen 3 235B \shortcite{qwen3-235} & \textbf{0.771} & \textbf{0.776} & \textbf{0.852} & \textbf{0.739} & \textbf{0.950} & \textbf{0.783} & \textbf{0.938} & \textbf{0.874} & \textbf{0.819} & \textbf{0.803} & 0.610 & 0.815 \\
DeepSeek V3 \shortcite{deepseek} & 0.739 & 0.705 & 0.820 & 0.698 & \textbf{0.950} & 0.743 & \textbf{0.938} & 0.863 & 0.794 & 0.774 & 0.603 & \textbf{0.818} \\
GPT-4o \shortcite{openai2024gpt4ocard} & 0.734 & 0.717 & 0.762 & 0.704 & 0.875 & 0.743 & 0.875 & 0.822 & 0.767 & 0.746 & \textbf{0.653} & \textbf{0.818} \\
Qwen3-Next 80B \shortcite{qwen3-235} & 0.732 & 0.720 & 0.803 & 0.700 & 0.925 & 0.739 & \textbf{0.938} & 0.855 & 0.782 & 0.744 & 0.590 & 0.793 \\
Sonnet 4 \shortcite{sonnet} & 0.726 & 0.756 & 0.713 & 0.693 & 0.875 & 0.770 & 0.812 & 0.798 & 0.787 & 0.750 & 0.613 & 0.771 \\
Llama 3.3 70B \shortcite{llama3} & 0.725 & 0.709 & 0.779 & 0.690 & 0.875 & 0.686 & 0.875 & 0.839 & 0.804 & 0.726 & 0.640 & 0.799 \\
OLMo 2 7B \shortcite{olmo2} & 0.608 & 0.626 & 0.730 & 0.568 & 0.750 & 0.566 & 0.812 & 0.727 & 0.662 & 0.636 & 0.450 & 0.702 \\
OLMo 2 32B \shortcite{olmo2} & 0.567 & 0.516 & 0.721 & 0.516 & 0.750 & 0.527 & 0.812 & 0.738 & 0.615 & 0.605 & 0.463 & 0.650 \\
Llama 4 Scout \shortcite{llama4} & 0.465 & 0.413 & 0.590 & 0.363 & 0.600 & 0.593 & 0.812 & 0.724 & 0.625 & 0.520 & 0.283 & 0.581 \\

\hline
\end{tabular}

\caption{Model accuracies (proportion correct) on the FinForge benchmark (5k samples). \textbf{Abbreviations:} Alt/RE = Alternative Investments \& Real Estate; Beh/Quant = Behavioral \& Quant Finance; Corp Fin \& Val = Corporate Finance \& Valuation; FinTech = FinTech \& Innovation; FAR = Financial Accounting \& Reporting; Ethics \& Gov = Financial Ethics \& Governance; Mkt \& Deriv = Markets \& Derivatives; Reg \& Comp = Regulation \& Compliance; Portf Mgmt = Investment \& Portfolio Management; Wealth Mgmt = Personal Finance \& Wealth Management; Pub/Intl Fin = Public \& International Finance.}
\label{tab:model_performance}

\end{table*}

\section{Results}
To demonstrate the pipeline's efficiency, the complete corpus construction and question-generation workflow was executed over seven days, encompassing taxonomy design, domain filtering, and content extraction. For this snapshot generation, the data curation stage yielded over \textbf{100{,}000} high-quality and verified financial documents spanning eleven well-defined subdomains. Collectively, these documents contained \textbf{143M tokens} of domain-specific text, providing a diverse foundation that integrates both quantitative and conceptual financial knowledge.

From this corpus, we sampled 10{,}000 documents for question generation via stratified random sampling across the 11 subdomains, ensuring proportional representation using \texttt{Gemini 2.5 Flash} as the synthesis model. This process produced \textbf{10{,}000} initial question–answer (Q/A) pairs. Subsequent automated filtering—combining rule-based quality control with LM-as-judge scoring for relevance, clarity, and factual accuracy—resulted in the refined \textbf{FinForge-5k} benchmark set of \textbf{5{,}000} Q/A pairs. The primary rejection reasons were insufficient self-containment (questions requiring external context), ambiguous answer choices, and misalignment between question difficulty and source material complexity. A stratified random sample of \textbf{500} Q/A pairs was additionally reviewed by domain experts (see Expert Evaluation).

\subsection{Model Benchmarking}
The validated \textbf{FinForge-5k} benchmark was used to evaluate a range of both open-source and closed-source language models (Table~\ref{tab:model_performance}). Performance was measured using a consistent multiple-choice setup, with accuracy defined as the proportion of correctly predicted answers across 5{,}000 question–answer pairs.

We organize the evaluation along two dimensions: model availability (proprietary vs. open-source) and scale (parameter count). Closed-source models such as \texttt{GPT-4o} and \texttt{Claude Sonnet 4} achieve accuracies of \textbf{73.4}\% and \textbf{72.6}\%, respectively. Notably, open-source models of the same generation, such as \texttt{Qwen-3-235B} and \texttt{DeepSeek v3.1}, demonstrate superior performance on the benchmark.

Additionally, we have categorized the evaluated models into three key groups to assess the influence of model scale on their performance against the benchmark. Surprisingly, mid-range models (32-110B) demonstrate performance comparable to large-scale proprietary models, with \texttt{Qwen3-Next-80B} exhibiting only a 5\% deficit relative to \texttt{Qwen-3-235B} and remaining within 1\% of the DeepSeek, GPT-4o, and Sonnet models. This suggests that sheer model scale is not the sole determinant of success in domain-specific financial reasoning, underscoring the utility of \textbf{FinForge-5k} in diagnosing reasoning capabilities beyond generalist benchmarks. We hypothesize that observed performance differences may reflect variations in training data composition or domain-specific optimization strategies; however, confirming these hypotheses would require access to training corpora and implementation details that are not publicly disclosed.

These results reveal a key insight for financial evaluation: neither state-of-the-art generalist performance nor model scale alone guarantees superior financial reasoning capabilities.

\subsection{Expert Evaluation}
To validate the efficacy of the pipeline's generation capability, three domain experts conducted a qualitative assessment on a 10\% stratified sample of the FinForge-5k snapshot. This expert review entailed verifying the clarity, self-containment, plausibility, and real-world relevance of the generated items.

The expert review validated the high quality and complexity of the generated questions, with 70\% of the 500 samples deemed clear, accurate, and relevant. Crucially, the remaining 30\% were typically not factually incorrect, but were flagged for ambiguity or missing contextual assumptions required for a definitive answer. This validation rate stands in significant contrast to the 100\% approval rate assigned by the automated LM-as-a-judge (Stage 5) for the identical sample. The 30-point discrepancy offers quantitative support for our conclusion regarding the limitations of the 'LM-as-a-judge,' indicating that it currently lacks the sophistication required to assess complex financial reasoning, demonstrating that expert oversight remains essential.

\subsection{Addressing Data Contamination and Circularity} Data contamination---where test data appears in training corpora---remains a significant concern for LLM evaluation \cite{survey2025contamination}. As a methodological sanity check, we evaluated the generator model (Gemini 2.5 Flash) and a model from the same family (Gemma 27B \citet{gemma3}) on the resulting benchmark. They achieved scores of 79.3\% and 74.0\%, respectively. Due to the inherent risk of data contamination from this circular evaluation, we explicitly omit the Gemma and Gemini models from our primary benchmark (Table~\ref{tab:model_performance}) to ensure a fair and rigorous assessment of other models.

\section{Discussion}
The FinForge-5k snapshot reveals systematic weaknesses in current model capabilities. Beyond the accuracy scores in \textbf{Table~\ref{tab:model_performance}}, we conduct a proportional error analysis to identify specific reasoning deficiencies.

Our analysis indicates that \textbf{\textit{Personal Finance \& Wealth Management}} and \textbf{\textit{Corporate Finance \& Valuation}} are notably challenging topics relative to others in the benchmark. Fundamentally, we characterize these domains as tasks requiring complex multi-constraint satisfaction—such as optimizing for tax liabilities, liquidity needs, and risk simultaneously—in contrast to traditional retrieval-based finance tasks. This contrasts with the models' comparatively high performance on Markets and Derivatives and Portfolio Management. This disparity suggests that despite exposure to substantial publicly available financial data during pretraining \cite{wu2023bloomberggpt}, models struggle to transfer such knowledge to scenarios requiring the integration of regulatory constraints with fundamental financial principles.

Conversely, across subjects, evidence indicates that models struggle most with quantitative questions, followed by counterfactual inquiries, aligning with the recent understanding of language model capabilities \cite{ahn2024llmmathreasoning,mirzadeh2024gsmsymbolic}. It is noteworthy that while the FinForge framework can generate multi-hop reasoning questions, for this snapshot they were frequently derived from distinct sections of single documents, representing a specific tier of reasoning difficulty.

Expert evaluation of incorrect quantitative answers in FinForge-5k identified two distinct failure modes. The first was conceptual: models applied incorrect financial methodologies, made flawed assumptions, or constructed inadequate logic (\eg miscalculating depreciation). The second was arithmetic: models recognized the correct steps but erred in calculation. This distinction is essential. Arithmetic failures can be mitigated by external tools \cite{schick2023toolformer,gao2023pal}, but conceptual failures reveal a fundamental deficiency: models are misinterpreting the intricate financial reasoning demanded by the prompt. Future studies should concentrate on addressing these conceptual misinterpretations.

The findings possess substantial real-world implications. The significant shortcomings in \textit{Personal Finance \& Wealth Management} highlight that the average user cannot currently rely on these models for important financial decisions \cite{canaihelppf,aregenaiagents}. This represents a significant gap that has previously been overlooked in benchmarks focused on conventional information retrieval.

This work contributes in two ways. We first identify and analyze specific, high-impact weaknesses in contemporary LMs concerning financial reasoning and personalization. We demonstrate that the FinForge framework is an effective method for generating nuanced, domain-specific benchmarks. This outlines a clear framework for guiding future advancements and focused enhancements to financial-sector models.

\section{Conclusion}
FinForge demonstrates that scalable, domain-grounded benchmark generation is achievable through a principled combination of expert oversight and controlled LM synthesis. By integrating verified financial sources with structured question generation and multi-stage validation, FinForge produces high-quality datasets that accurately reflect the depth of reasoning and quantitative rigor demanded in economic analysis. Evaluations using the FinForge-5k snapshot reveal specific, high-impact weaknesses in state-of-the-art models, particularly a tendency to fail at conceptual reasoning rather than simple arithmetic. This validation gap reinforces the importance of expert oversight in complex domains. 
Beyond the immediate utility of the FinForge-5k snapshot, the framework offers a generalizable methodology for creating transparent, extendable evaluation pipelines in other specialized domains. Future work will focus on expanding the corpus to additional subfields, integrating temporal and dynamic data to assess model recency, and establishing continual-learning benchmarks that evolve alongside financial markets. Ultimately, FinForge lays the groundwork for systematic, reproducible, and evolving evaluation of LMs in finance and other expert-driven disciplines.

\section{Limitations}
\label{sec:limitations}
The primary limitation of the study is the reliance on Gemini 2.5 Flash for both question generation and evaluation. Manual verification revealed that while questions were challenging, they often lacked the contextual assumptions needed for a definitive answer. The ambiguity, likely arising from the generator's speed optimization, poses a risk of leading assessed LMs to make incorrect assumptions, thereby compromising the reliability of assessments.

Gemini 2.5 Flash's design, which prioritizes speed, results in a capabilities mismatch when evaluating complex reasoning models. This likely led to the evaluator lacking the sophistication necessary to accurately assess advanced outputs, which may have distorted performance metrics. The boolean validator operated as a ``black box.'' The system effectively filtered out irrelevant questions; however, its lack of transparency obstructed the analysis of failed questions, thereby impeding the improvement of the generation pipeline.

Additionally, data contamination and circular dependency remain concerns when evaluating models from the same family as the generator. The present study omits the performance of Gemini and Gemma models from the primary results to mitigate this issue, but future work should develop more robust strategies for isolating generator influence from evaluation outcomes.




\bibliography{references}

@misc{openai2023gpt4,
      title={GPT-4 Technical Report},
      author={OpenAI},
      year={2023},
      eprint={2303.08774},
      archivePrefix={arXiv},
      primaryClass={cs.CL},
      url={https://arxiv.org/abs/2303.08774},
}

@misc{openai2024gpt4ocard,
      title={GPT-4o System Card},
      author={OpenAI team and others},
      year={2024},
      eprint={2410.21276},
      archivePrefix={arXiv},
      primaryClass={cs.CL},
      url={https://arxiv.org/abs/2410.21276},
}

@techreport{sonnet,
  author    = {{Anthropic}},
  title     = {System Card: Claude Opus 4 \& Claude Sonnet 4},
  institution = {Anthropic},
  year      = {2025},
  month     = {May},
  url       = {https://www-cdn.anthropic.com/4263b940cabb546aa0e3283f35b686f4f3b2ff47.pdf}
}

@misc{deepseek,
      title={DeepSeek-V3 Technical Report},
      author={DeepSeek-AI},
      year={2024},
      eprint={2412.19437},
      archivePrefix={arXiv},
      primaryClass={cs.CL},
      url={https://arxiv.org/abs/2412.19437},
}

@misc{qwen3-235,
      title={Qwen3 Technical Report},
      author={Qwen Team},
      year={2025},
      eprint={2505.09388},
      archivePrefix={arXiv},
      primaryClass={cs.CL},
      url={https://arxiv.org/abs/2505.09388},
}

@misc{llama3,
      title={The Llama 3 Herd of Models},
      author={{Meta AI}},
      year={2024},
      eprint={2407.21783},
      archivePrefix={arXiv},
      primaryClass={cs.AI},
      url={https://arxiv.org/abs/2407.21783},
}

@misc{llama4,
  author    = {{Meta AI}},
  title     = {The Llama 4 herd: The beginning of a new era of natively multimodal AI innovation},
  year      = {2025},
  month     = {April},
  url       = {https://ai.meta.com/blog/llama-4-multimodal-intelligence/}
}

@misc{gemini,
      title={Gemini 2.5: Pushing the Frontier with Advanced Reasoning, Multimodality, Long Context, and Next Generation Agentic Capabilities},
      author ={Gemini Team and others},
      year={2025},
      eprint={2507.06261},
      archivePrefix={arXiv},
      primaryClass={cs.CL},
      url={https://arxiv.org/abs/2507.06261},
}

@misc{gemma3,
      title={Gemma 3 Technical Report},
      author={Gemma Team and others},
      year={2025},
      eprint={2503.19786},
      archivePrefix={arXiv},
      primaryClass={cs.CL},
      url={https://arxiv.org/abs/2503.19786},
}

@article{olmo2,
      title={2 OLMo 2 Furious},
      author={Team OLMo and others},
      year={2024},
      eprint={2501.00656},
      archivePrefix={arXiv},
      primaryClass={cs.CL},
      url={https://arxiv.org/abs/2501.00656},
}

@misc{wu2023bloomberggpt,
  title={BloombergGPT: A Large Language Model for Finance},
  author={Wu, Shijie and Irsoy, Ozan and Lu, Steven and Dabravolski, Vadim and Dredze, Mark and Gehrmann, Sebastian and Kambadur, Prabhanjan and Rosenberg, David and Mann, Gideon},
  year={2023},
  eprint={2303.17564},
  archivePrefix={arXiv},
  primaryClass={cs.CL},
  url={https://arxiv.org/abs/2303.17564}
}

@article{bommasani2022opportunities,
  author       = {Rishi Bommasani and
                  Drew A. Hudson and
                  Ehsan Adeli and
                  Russ B. Altman and
                  Simran Arora and
                  Sydney von Arx and
                  Michael S. Bernstein and
                  Jeannette Bohg and
                  Antoine Bosselut and
                  Emma Brunskill and
                  Erik Brynjolfsson and
                  Shyamal Buch and
                  Dallas Card and
                  Rodrigo Castellon and
                  Niladri S. Chatterji and
                  Annie S. Chen and
                  Kathleen Creel and
                  Jared Quincy Davis and
                  Dorottya Demszky and
                  Chris Donahue and
                  Moussa Doumbouya and
                  Esin Durmus and
                  Stefano Ermon and
                  John Etchemendy and
                  Kawin Ethayarajh and
                  Li Fei{-}Fei and
                  Chelsea Finn and
                  Trevor Gale and
                  Lauren E. Gillespie and
                  Karan Goel and
                  Noah D. Goodman and
                  Shelby Grossman and
                  Neel Guha and
                  Tatsunori Hashimoto and
                  Peter Henderson and
                  John Hewitt and
                  Daniel E. Ho and
                  Jenny Hong and
                  Kyle Hsu and
                  Jing Huang and
                  Thomas Icard and
                  Saahil Jain and
                  Dan Jurafsky and
                  Pratyusha Kalluri and
                  Siddharth Karamcheti and
                  Geoff Keeling and
                  Fereshte Khani and
                  Omar Khattab and
                  Pang Wei Koh and
                  Mark S. Krass and
                  Ranjay Krishna and
                  Rohith Kuditipudi and
                  et al.},
  title        = {On the Opportunities and Risks of Foundation Models},
  journal      = {CoRR},
  volume       = {abs/2108.07258},
  year         = {2021},
  url          = {https://arxiv.org/abs/2108.07258},
  eprinttype    = {arXiv},
  eprint       = {2108.07258},
  bibsource    = {dblp computer science bibliography, https://dblp.org}
}

@article{hendryckstest2021,
  title={Measuring Massive Multitask Language Understanding},
  author={Dan Hendrycks and Collin Burns and Steven Basart and Andy Zou and Mantas Mazeika and Dawn Song and Jacob Steinhardt},
  journal={International Conference on Learning Representations},
  year={2021}
}

@article{allenai:arc,
  author = {Peter Clark and Isaac Cowhey and Oren Etzioni and Tushar Khot and Ashish Sabharwal and Carissa Schoenick and Oyvind Tafjord},
  title = {Think you have Solved Question Answering? Try ARC, the AI2 Reasoning Challenge},
  year = {2018},
  url = {https://allenai.org/data/arc}
}

@misc{arcbench,
      title={ARC Prize 2024: Technical Report},
      author={Francois Chollet and Mike Knoop and Gregory Kamradt and Bryan Landers},
      year={2025},
      eprint={2412.04604},
      archivePrefix={arXiv},
      primaryClass={cs.AI},
      url={https://arxiv.org/abs/2412.04604},
}

@misc{bigbench,
      title={Beyond the Imitation Game: Quantifying and extrapolating the capabilities of language models},
      author={Aarohi Srivastava and Abhinav Rastogi and Abhishek Rao and Abu Awal Md Shoeb and Abubakar Abid and Adam Fisch and Adam R. Brown and Adam Santoro and Aditya Gupta and others},
      year={2023},
      eprint={2206.04615},
      archivePrefix={arXiv},
      primaryClass={cs.CL},
      url={https://arxiv.org/abs/2206.04615},
}

@inproceedings{koncelkedziorski2024bizbench,
  title={BizBench: A Quantitative Reasoning Benchmark for Business and Finance},
  author={Koncel-Kedziorski, Rik and others},
  booktitle={Proceedings of the 62nd Annual Meeting of the Association for Computational Linguistics},
  year={2024},
  publisher={Association for Computational Linguistics},
  url={https://aclanthology.org/2024.acl-long.452/}
}

@misc{butt2024benchagents,
  title={BenchAgents: Multi-Agent Systems for Structured Benchmark Creation},
  author={Butt, Safwan and Chandrasekaran, Dhanuja and Moorthy, Sai Pragna and Agrawal, Raunak and others},
  year={2024},
  eprint={2410.22584},
  archivePrefix={arXiv},
  primaryClass={cs.CL},
  url={https://arxiv.org/abs/2410.22584}
}

@inproceedings{deng2024investigating,
  title={Investigating Data Contamination in Modern Benchmarks for Large Language Models},
  author={Deng, Chunyuan and Zhao, Yilun and Tang, Xiangru and Gerstein, Mark and Cohan, Arman},
  booktitle={Proceedings of the 2024 Conference of the North American Chapter of the Association for Computational Linguistics},
  year={2024},
  publisher={Association for Computational Linguistics},
  url={https://aclanthology.org/2024.naacl-long.482/},
  doi={10.18653/v1/2024.naacl-long.482}
}

@misc{benchmark2024contamination,
      title={Benchmark Data Contamination of Large Language Models: A Survey},
      author={Cheng Xu and Shuhao Guan and Derek Greene and M-Tahar Kechadi},
      year={2024},
      eprint={2406.04244},
      archivePrefix={arXiv},
      primaryClass={cs.CL},
      url={https://arxiv.org/abs/2406.04244},
}

@misc{xu2024benchmarking,
      title={Benchmarking Benchmark Leakage in Large Language Models},
      author={Ruijie Xu and Zengzhi Wang and Run-Ze Fan and Pengfei Liu},
      year={2024},
      eprint={2404.18824},
      archivePrefix={arXiv},
      primaryClass={cs.CL},
      url={https://arxiv.org/abs/2404.18824},
}

@misc{survey2025contamination,
      title={A Survey on Data Contamination for Large Language Models},
      author={Yuxing Cheng and Yi Chang and Yuan Wu},
      year={2025},
      eprint={2502.14425},
      archivePrefix={arXiv},
      primaryClass={cs.CL},
      url={https://arxiv.org/abs/2502.14425},
}

@inproceedings{li2024latesteval,
  title={LatestEval: Addressing Data Contamination in Language Model Evaluation through Dynamic and Time-Sensitive Test Construction},
  author={Li, Yucheng and Guerin, Frank and Lin, Chenghua},
  booktitle={Proceedings of the AAAI Conference on Artificial Intelligence},
  year={2024},
  url={https://arxiv.org/abs/2312.12343}
}

@misc{savaal,
      title={Savaal: Scalable Concept-Driven Question Generation to Enhance Human Learning},
      author={Kimia Noorbakhsh and Joseph Chandler and Pantea Karimi and Mohammad Alizadeh and Hari Balakrishnan},
      year={2025},
      eprint={2502.12477},
      archivePrefix={arXiv},
      primaryClass={cs.CL},
      url={https://arxiv.org/abs/2502.12477},
}

@inproceedings{pfqs,
    title = "Planning First, Question Second: An {LLM}-Guided Method for Controllable Question Generation",
    author = "Li, Kunze  and
      Zhang, Yu",
    editor = "Ku, Lun-Wei  and
      Martins, Andre  and
      Srikumar, Vivek",
    booktitle = "Findings of the Association for Computational Linguistics: ACL 2024",
    month = aug,
    year = "2024",
    address = "Bangkok, Thailand",
    publisher = "Association for Computational Linguistics",
    url = "https://aclanthology.org/2024.findings-acl.280/",
    doi = "10.18653/v1/2024.findings-acl.280",
    pages = "4715--4729"
}

@article{Das2021AutomaticQG,
  author  = {Das, B. and Majumder, M. and Phadikar, S. and Sekh, A.A.},
  title   = {Automatic question generation and answer assessment: A survey},
  journal = {Research and Practice in Technology Enhanced Learning},
  volume  = {16},
  number  = {5},
  pages   = {1--15},
  year    = {2021},
  doi     = {10.1186/s41039-021-00151-1}
}

@article{Guo2024SurveyNQG,
  title={A Survey on Neural Question Generation: Methods, Applications, and Prospects},
  author={Guo, Shasha and Liao, Lizi and Li, Cuiping and Chua, Tat-Seng},
  journal={arXiv preprint arXiv:2402.18267},
  year={2024}
}

@inproceedings{chen2022convfinqa,
  title={ConvFinQA: Exploring the Chain of Numerical Reasoning in Conversational Finance Question Answering},
  author={Chen, Zhiyu and Li, Shiyang and Smiley, Charese and Ma, Zhiqiang and Shah, Sameena and Wang, William Yang},
  booktitle={Proceedings of the 2022 Conference on Empirical Methods in Natural Language Processing},
  pages={6279--6292},
  year={2022},
  address={Abu Dhabi, United Arab Emirates},
  publisher={Association for Computational Linguistics},
  url={https://aclanthology.org/2022.emnlp-main.421},
  doi={10.18653/v1/2022.emnlp-main.421}
}

@inproceedings{chen2021finqa,
  title={FinQA: A Dataset of Numerical Reasoning over Financial Data},
  author={Chen, Zhiyu and Chen, Wenhu and Smiley, Charese and Shah, Sameena and Borova, Iana and Langdon, Dylan and Moussa, Reema and Beane, Matt and Huang, Ting-Hao and Routledge, Bryan R and others},
  booktitle={Proceedings of the 2021 Conference on Empirical Methods in Natural Language Processing},
  pages={3697--3711},
  year={2021}
}

@inproceedings{zhu-etal-2021-tat-qa,
    title = "{TAT-QA}: A Question Answering Benchmark on a Hybrid of Tabular and Textual Content in Finance",
    author = "Zhu, Fengbin and Lei, Wenqiang and Huang, Youcheng and Wang, Chao and Zhang, Shuo and Lv, Jiancheng and Feng, Fuli and Chua, Tat-Seng",
    booktitle = "Proceedings of the 59th Annual Meeting of the Association for Computational Linguistics and the 11th International Joint Conference on Natural Language Processing (Volume 1: Long Papers)",
    month = aug,
    year = "2021",
    address = "Online",
    publisher = "Association for Computational Linguistics",
    url = "https://aclanthology.org/2021.acl-long.254",
    doi = "10.18653/v1/2021.acl-long.254",
    pages = "3277--3287",
}

@inproceedings{Maia2018FiQA,
  author    = {Macedo Maia, S. and Handschuh, Siegfried and Freitas, Andr{\'e} and Davis, Brian and McDermott, Ross and Zarrouk, Manel and Balahur, Alexandra},
  title     = {{WWW'18 Open Challenge: Financial Opinion Mining and Question Answering}},
  booktitle = {{WWW '18 Companion: The 2018 Web Conference Companion}},
  pages     = {1941--1942},
  year      = {2018},
  address   = {Lyon, France},
  publisher = {ACM},
  isbn      = {9781450356404},
  doi       = {10.1145/3184558.3192301},
}

@inproceedings{chen-etal-2024-fintextqa,
    title = "{F}in{T}ext{QA}: A Dataset for Long-form Financial Question Answering",
    author = "Chen, Jian and Zhou, Peilin and Hua, Yining and Xin, Loh and Chen, Kehui and Li, Ziyuan and Zhu, Bing and Liang, Junwei",
    editor = "Ku, Lun-Wei and Martins, Andre and Srikumar, Vivek",
    booktitle = "Proceedings of the 62nd Annual Meeting of the Association for Computational Linguistics (Volume 1: Long Papers)",
    month = aug,
    year = "2024",
    address = "Bangkok, Thailand",
    publisher = "Association for Computational Linguistics",
    url = "https://aclanthology.org/2024.acl-long.328/",
    doi = "10.18653/v1/2024.acl-long.328",
}

@article{Islam2023FinanceBenchAN,
  title={FinanceBench: A New Benchmark for Financial Question Answering},
  author={Islam, Pranab and Kannappan, Anand and Kiela, Douwe and Qian, Rebecca and Scherrer, Nino and Vidgen, Bertie},
  journal={arXiv preprint arXiv:2311.11944},
  year={2023}
}

@misc{cfa2024evaluation,
      title={Evaluating Large Language Models for Financial Reasoning: A CFA-Based Benchmark Study},
      author={Xuan Yao and Qianteng Wang and Xinbo Liu and Ke-Wei Huang},
      year={2024},
      eprint={2509.04468},
      archivePrefix={arXiv},
      primaryClass={cs.CL},
      url={https://arxiv.org/abs/2509.04468},
}

@misc{cfa2025advanced,
      title={Advanced Financial Reasoning at Scale: A Comprehensive Evaluation of Large Language Models on CFA Level III},
      author={Pranam Shetty and Abhisek Upadhayaya and Parth Mitesh Shah and Srikanth Jagabathula and Shilpi Nayak and Anna Joo Fee},
      year={2025},
      eprint={2507.02954},
      archivePrefix={arXiv},
      primaryClass={cs.CL},
      url={https://arxiv.org/abs/2507.02954},
}

@misc{finllms2024survey,
      title={A Survey of Large Language Models for Financial Applications: Progress, Prospects and Challenges},
      author={Yuqi Nie and Yaxuan Kong and Xiaowen Dong and John M. Mulvey and H. Vincent Poor and Qingsong Wen and Stefan Zohren},
      year={2024},
      eprint={2406.11903},
      archivePrefix={arXiv},
      primaryClass={cs.CL},
      url={https://arxiv.org/abs/2406.11903},
}

@article{dllf2024survey,
  title={Deep Learning in Finance: A Survey of Applications and Techniques},
  author={Kamalov, Firuz and Gurrib, Ikhlaas and Elmazi, Rajula and Moussa, Said},
  journal={AI},
  volume={5},
  number={4},
  pages={101},
  year={2024},
  publisher={MDPI},
  url={https://www.mdpi.com/2673-2688/5/4/101}
}

@article{ai2022finance,
  title={AI in Finance: Challenges, Techniques, and Opportunities},
  author={Cao, Longbing},
  journal={ACM Computing Surveys},
  volume={55},
  number={3},
  pages={1--38},
  year={2022},
  publisher={ACM},
  doi={10.1145/3502289}
}

@article{ai2025integration,
  title={AI integration in financial services: a systematic review of trends and regulatory challenges},
  author={Vukovi{\'c}, Darko B. and Dekpo-Adza, Senanu and Matovi{\'c}, Stefana},
  journal={Humanities and Social Sciences Communications},
  volume={12},
  year={2025},
  publisher={Nature},
  url={https://www.nature.com/articles/s41599-025-04850-8}
}

@article{biblio2024ai,
  title={Artificial Intelligence and Finance: A bibliometric review on the Trends, Influences, and Research Directions},
  author={Alzoubi, Mohammad and Alkhateeb, Ahmad and others},
  journal={F1000Research},
  volume={14},
  number={122},
  year={2025},
  url={https://pmc.ncbi.nlm.nih.gov/articles/PMC11795023/}
}

@article{canaihelppf,
   title={Can AI help with your personal finances?},
   ISSN={1466-4283},
   url={http://dx.doi.org/10.1080/00036846.2025.2450384},
   DOI={10.1080/00036846.2025.2450384},
   journal={Applied Economics},
   publisher={Informa UK Limited},
   author={Hean, Oudom and Saha, Utsha and Saha, Binita},
   year={2025},
   month=jan, pages={1–9} }

@misc{aregenaiagents,
      title={Are Generative AI Agents Effective Personalized Financial Advisors?},
      author={Takehiro Takayanagi and Kiyoshi Izumi and Javier Sanz-Cruzado and Richard McCreadie and Iadh Ounis},
      year={2025},
      eprint={2504.05862},
      archivePrefix={arXiv},
      primaryClass={cs.AI},
      url={https://arxiv.org/abs/2504.05862},
}

@misc{schick2023toolformer,
  title={Toolformer: Language Models Can Teach Themselves to Use Tools},
  author={Schick, Timo and Dwivedi-Yu, Jane and Dess{\`i}, Roberto and Raileanu, Roberta and Lomeli, Maria and Zettlemoyer, Luke and Cancedda, Nicola and Scialom, Thomas},
  year={2023},
  eprint={2302.04761},
  archivePrefix={arXiv},
  primaryClass={cs.CL},
  url={https://arxiv.org/abs/2302.04761}
}

@misc{gao2023pal,
  title={PAL: Program-aided Language Models},
  author={Gao, Luyu and Madaan, Aman and Zhou, Shuyan and Alon, Uri and Liu, Pengfei and Yang, Yiming and Callan, Jamie and Neubig, Graham},
  year={2023},
  eprint={2211.10435},
  archivePrefix={arXiv},
  primaryClass={cs.CL},
  url={https://arxiv.org/abs/2211.10435}
}

@inproceedings{ahn2024llmmathreasoning,
  title={Large Language Models for Mathematical Reasoning: Progresses and Challenges},
  author={Ahn, Janice and Verma, Rishu and Lou, Renze and Liu, Di and Zhang, Rui and Yin, Wenpeng},
  booktitle={Proceedings of the 18th Conference of the European Chapter of the Association for Computational Linguistics: Student Research Workshop},
  pages={225--237},
  year={2024},
  publisher={Association for Computational Linguistics},
  url={https://arxiv.org/abs/2402.00157}
}

@inproceedings{mirzadeh2024gsmsymbolic,
  title={GSM-Symbolic: Understanding the Limitations of Mathematical Reasoning in Large Language Models},
  author={Mirzadeh, Iman and Alizadeh, Keivan and Shahber, Hooman and Tuzel, Oncel and Benber, Samy and Farajtabar, Mehrdad},
  booktitle={International Conference on Learning Representations},
  year={2025},
  url={https://arxiv.org/abs/2410.05229}
}

@misc{long2024llmsyntheticsurvey,
  title={On LLMs-Driven Synthetic Data Generation, Curation, and Evaluation: A Survey},
  author={Long, Lin and others},
  year={2024},
  eprint={2406.15126},
  archivePrefix={arXiv},
  primaryClass={cs.CL},
  url={https://arxiv.org/abs/2406.15126}
}

\end{document}